\documentclass{llncs}
\pdfoutput=1
\usepackage{xspace}
\usepackage{xcolor}
\usepackage{url}
\usepackage{longtable}
\usepackage[utf8]{inputenc}
\usepackage{pdfpages}

\title{System Descriptions of the First International Competition on Computational Models of Argumentation (ICCMA'15)}
\author{Matthias Thimm, Serena Villata (Editors)}

\begin{document}
\noindent \LARGE Matthias Thimm, Serena Villata (Eds.)

\bigskip

\bigskip

\noindent \Huge System Descriptions of the \\First International Competition on\\ Computational Models of \\ Argumentation (ICCMA'15)

\vfill
\large October 2015
\frontmatter

\normalsize
\pagestyle{plain}

\pagenumbering{Roman}

\chapter*{Preface}
The objectives of the International Competition on Computational Models of Argumentation (ICCMA) are to provide a forum for empirical comparison of  solvers, to highlight challenges to the community, to propose new directions for research and to provide a core of common benchmark problems and a representation formalism that can aid in the comparison and evaluation of solvers.

The First International Competition on Computational Models of Argumentation (ICCMA'15) 
has been conducted in the first half of 2015 and focused on reasoning tasks in abstract 
argumentation frameworks. Submitted solvers were tested on several artificially 
generated argumentation frameworks in terms of correctness and performance. More precisely, solvers were evaluated based on their 
performance in solving the following computational tasks:
\begin{enumerate}
\item Given an abstract argumentation framework, determine some extension
\item Given an abstract argumentation framework, determine all extensions
\item Given an abstract argumentation framework and some argument, decide whether the given 
argument is credulously inferred
\item Given an abstract argumentation framework and some argument, decide whether the given 
argument is skeptically inferred
\end{enumerate}
The above computational tasks were to be solved with respect to the following standard 
semantics:
\begin{enumerate}
\item Complete Semantics
\item Preferred Semantics
\item Grounded Semantics
\item Stable Semantics
\end{enumerate}
Developers of solvers could provide support for a subset of the above 
computational tasks and/or semantics were also welcomed to support further semantics. 

This volume contains the system description of the 18 solvers submitted to the competition and therefore gives an overview on state-of-the-art of computational approaches to abstract argumentation problems. Further information on the results of the competition and the performance of the individual solvers can be found on at \url{http://argumentationcompetition.org/2015/}.

\vspace{1cm}

\begin{flushright}\noindent
October 2015\hfill Matthias Thimm, Serena Villata\\
General Chairs
\end{flushright}

\newpage

\section*{Organization}

\subsection*{General Chairs}
\begin{tabular}{@{}p{4cm}@{}p{8.2cm}@{}}
Matthias Thimm & University of Koblenz-Landau, Germany\\
Serena Villata & INRIA Sophia Antipolis, France
\end{tabular}
\subsection*{Steering Committee}
\begin{tabular}{@{}p{4cm}@{}p{8.2cm}@{}}
Nir Oren & University of Aberdeen, UK (\textit{President})\\
Hannes Strass & Leipzig University, Germany (\textit{Vice-President})\\
Mauro Vallati & University of Huddersfield, UK (\textit{Secretary})\\
Federico Cerutti & University of Aberdeen, UK\\
Matthias Thimm & University of Koblenz-Landau, Germany\\
Serena Villata &INRIA Sophia Antipolis, France
\end{tabular}

\tableofcontents
\mainmatter

\addcontentsline{toc}{chapter}{\normalfont \textit{Florian Brons}\\ LabSAT-Solver: Utilizing Caminada's Labelling Approach as a Boolean Satisfiability Problem}
\includepdf[pages=-,pagecommand={\thispagestyle{plain}}]{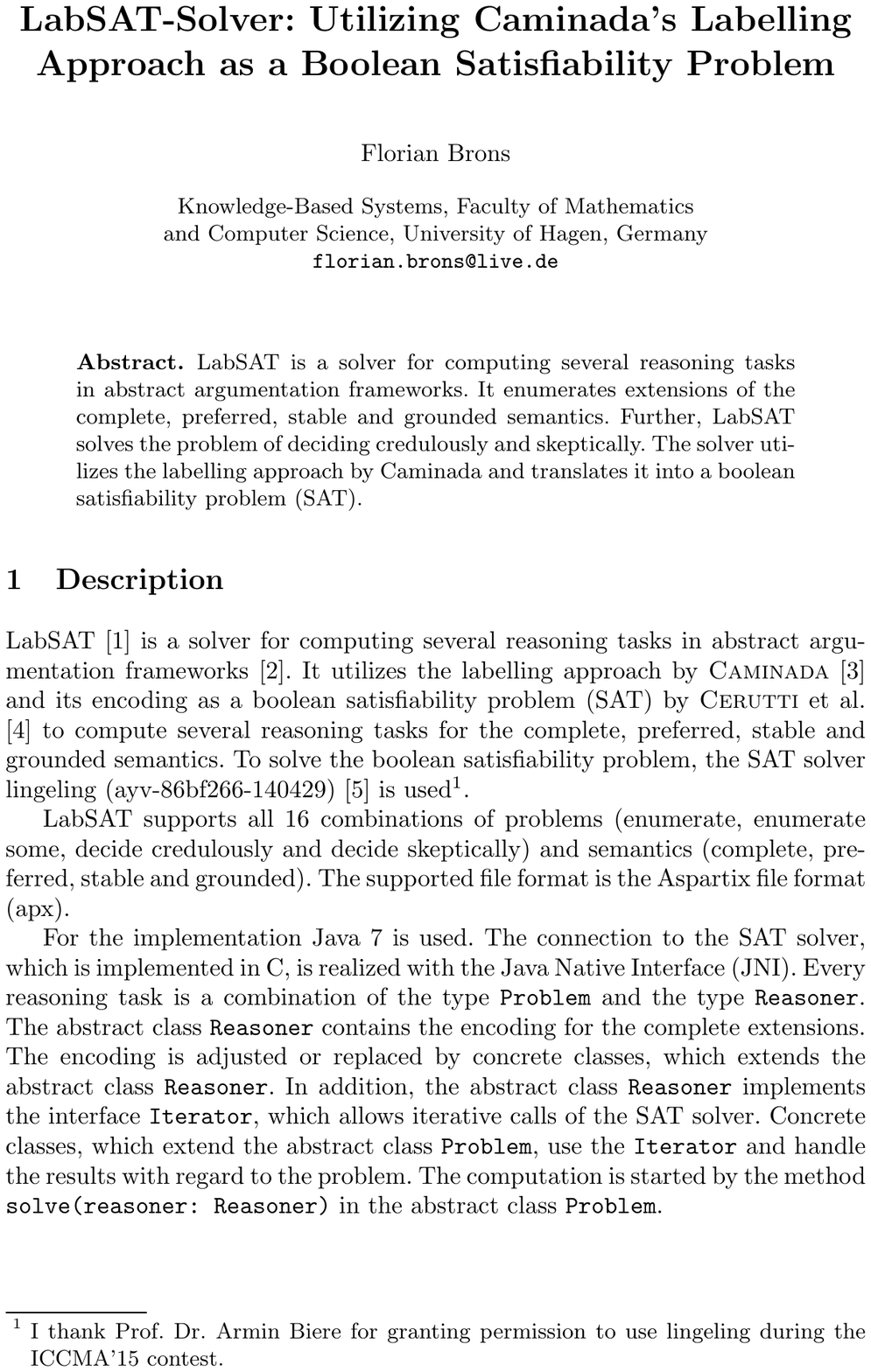}
\addcontentsline{toc}{chapter}{\normalfont \textit{Federico Cerutti, Mauro Vallati,Massimiliano Giacomin}\\ ArgSemSAT-1.0: Exploiting SAT Solvers in Abstract Argumentation}
\includepdf[pages=-,pagecommand={\thispagestyle{plain}}]{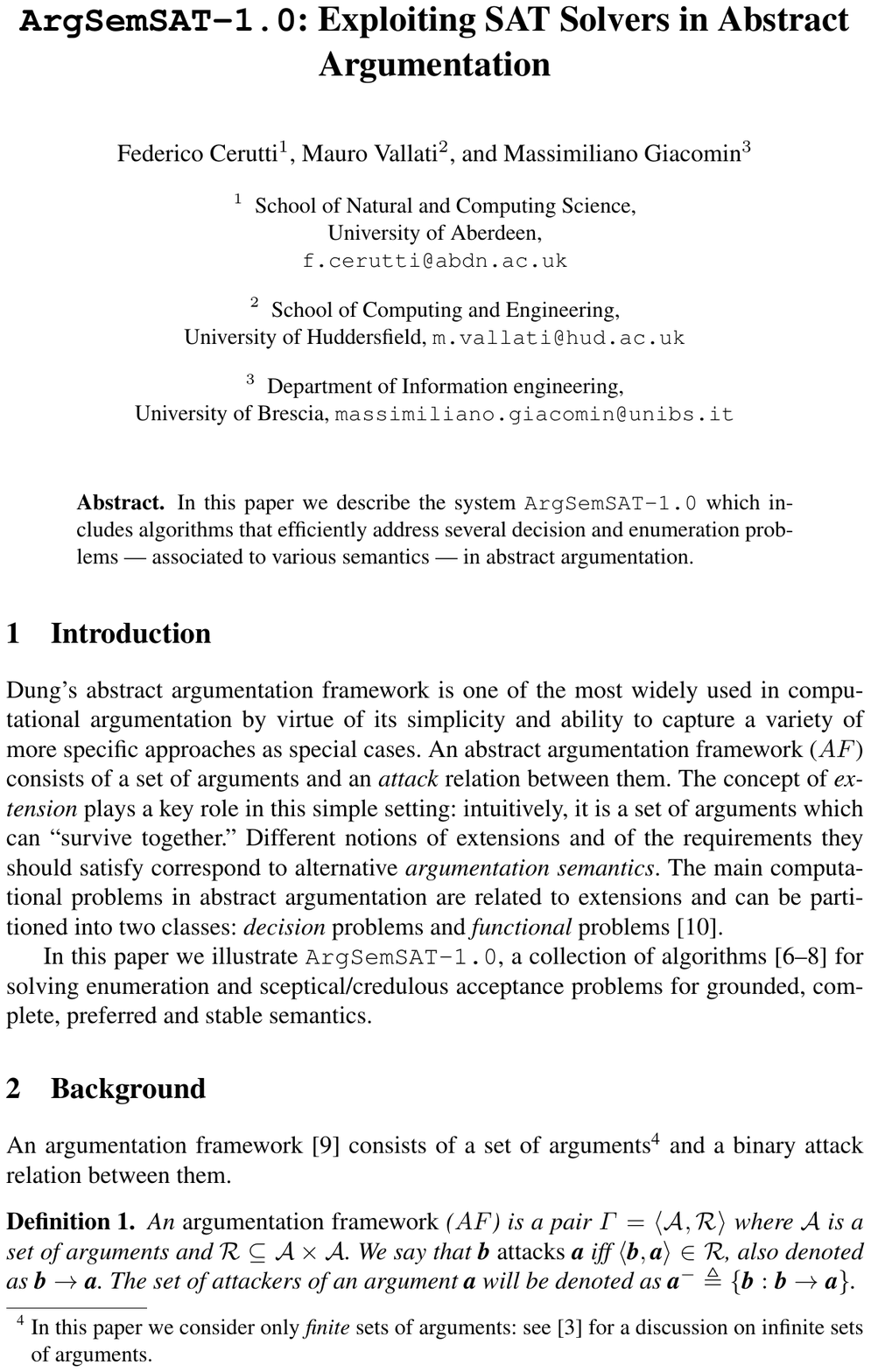}
\addcontentsline{toc}{chapter}{\normalfont \textit{Samer Nofal, Katie Atkinson, Paul E. Dunne}\\ ArgTools: a backtracking-based solver for abstract argumentation}
\includepdf[pages=-,pagecommand={\thispagestyle{plain}}]{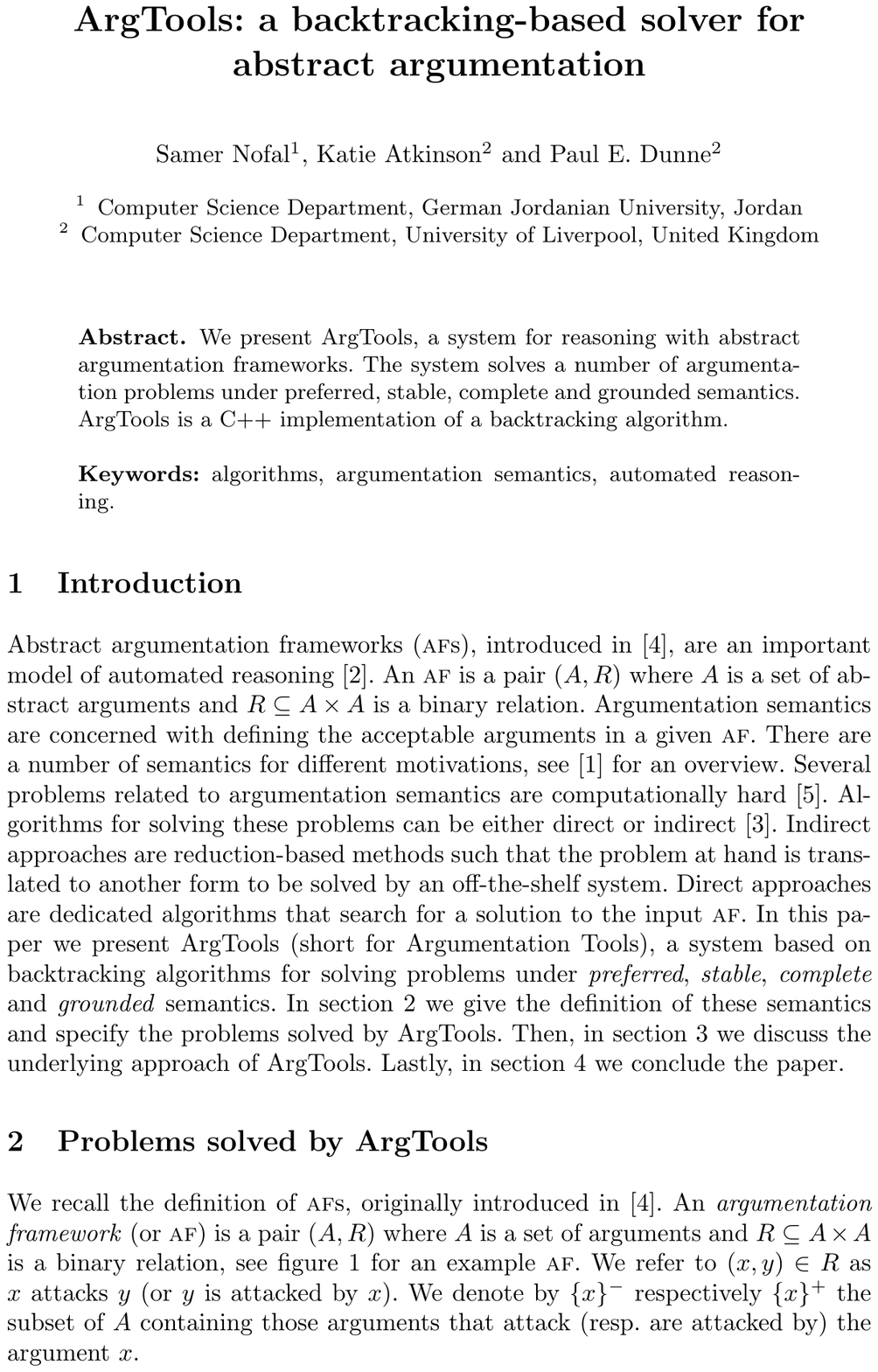}
\addcontentsline{toc}{chapter}{\normalfont \textit{Wolfgang Dvo\v{r}\'ak, Matti J\"arvisalo, Johannes Peter Wallner, Stefan Woltran}\\ CEGARTIX v0.4: A SAT-Based Counter-Example Guided Argumentation Reasoning Tool}
\includepdf[pages=-,pagecommand={\thispagestyle{plain}}]{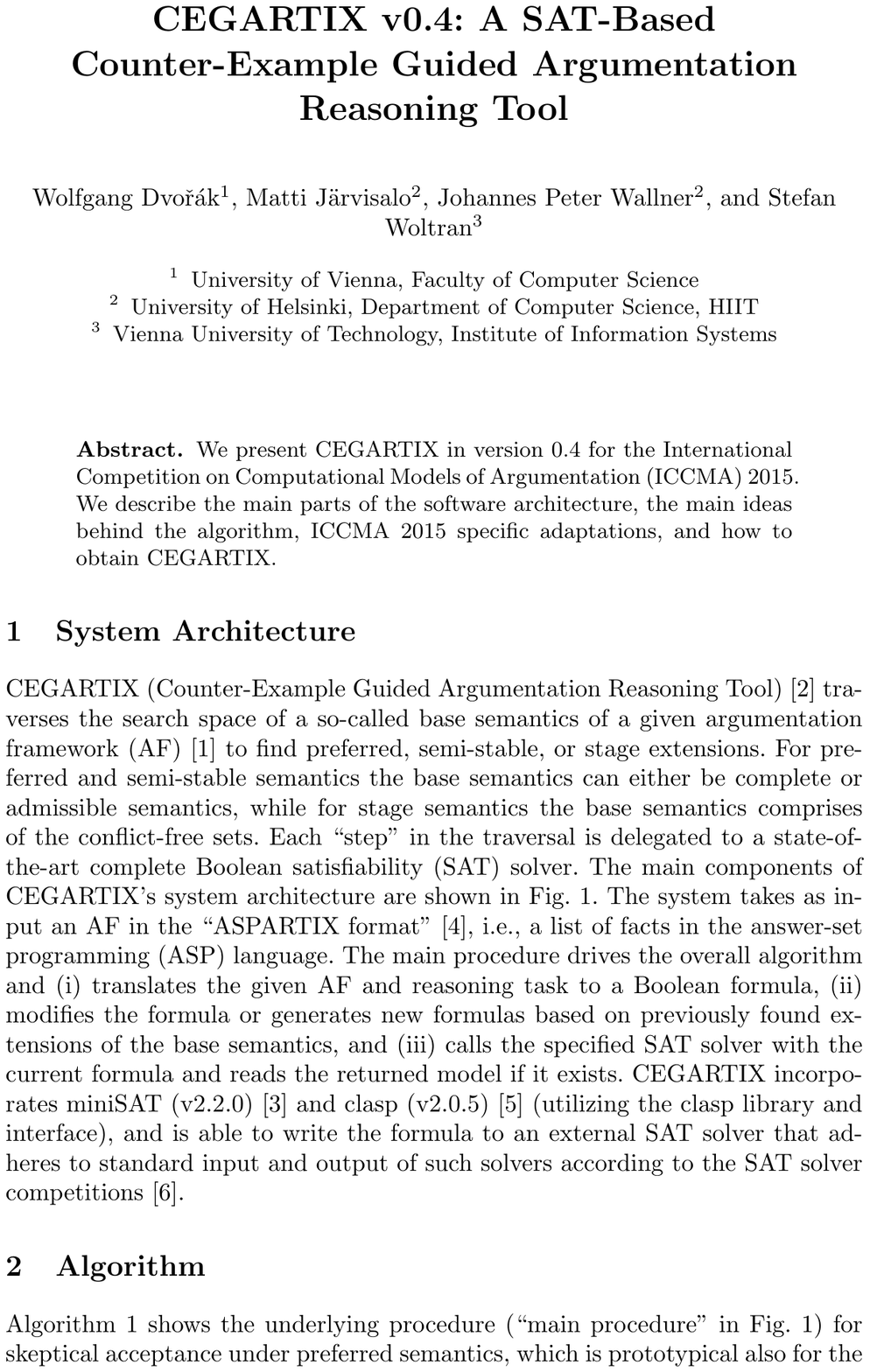}
\addcontentsline{toc}{chapter}{\normalfont \textit{Bas van Gijzel}\\ Dungell: A reference implementation of Dung's argumentation frameworks in Haskell}
\includepdf[pages=-,pagecommand={\thispagestyle{plain}}]{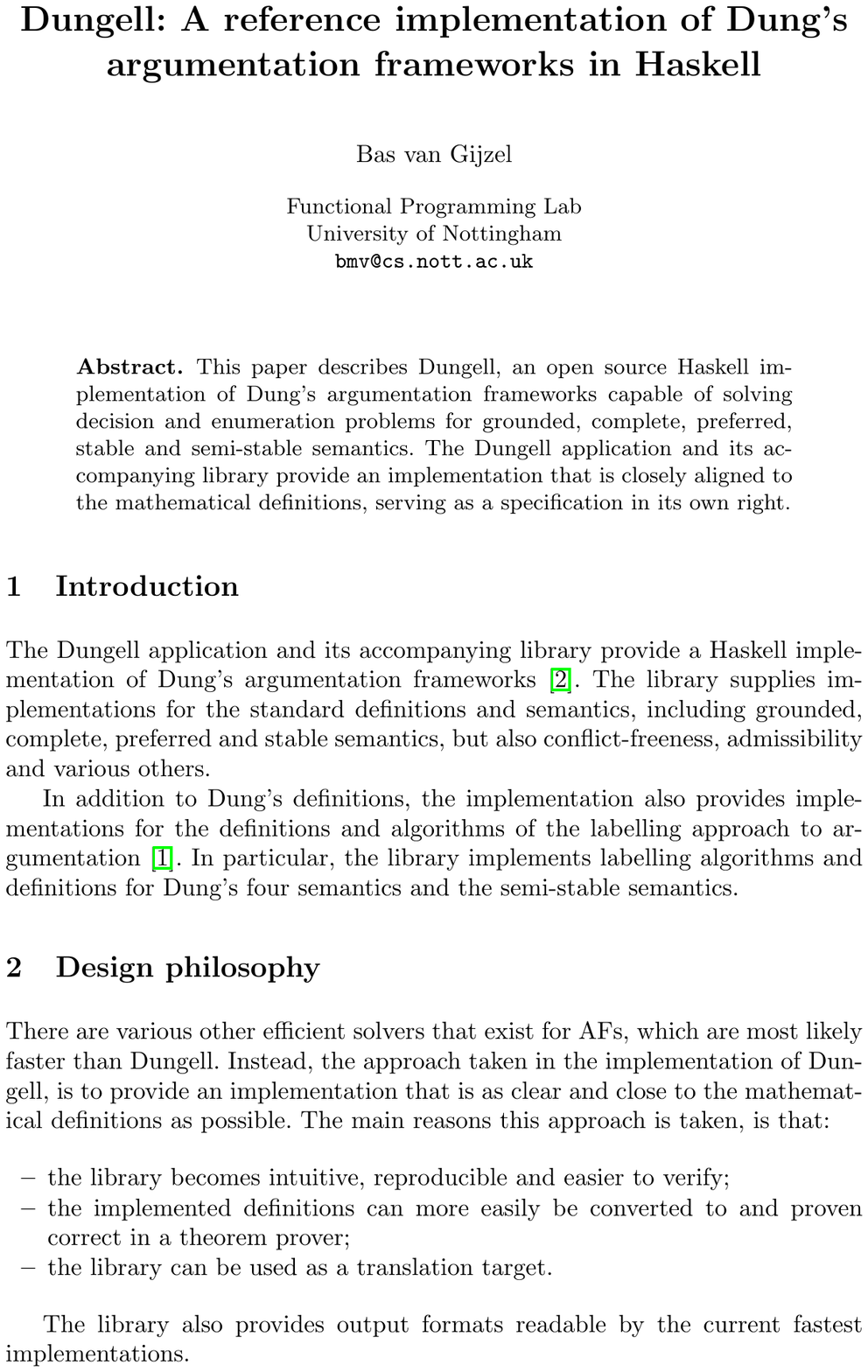}
\addcontentsline{toc}{chapter}{\normalfont \textit{Qianle Guo, Beishui Liao}\\ZJU-ARG: A Decomposition-Based Solver for Abstract Argumentation}
\includepdf[pages=-,pagecommand={\thispagestyle{plain}}]{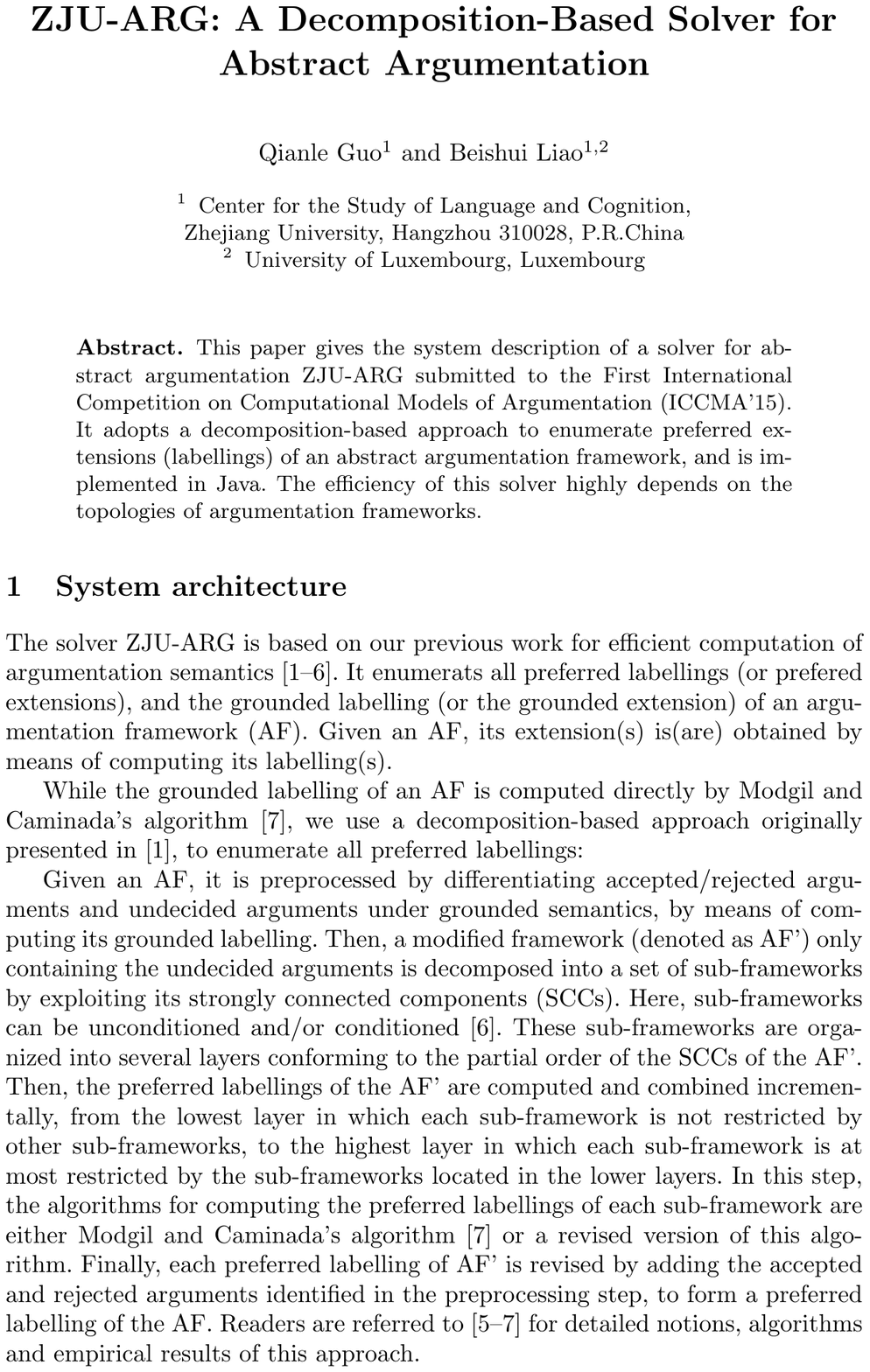}
\addcontentsline{toc}{chapter}{\normalfont \textit{Alessandro Ronca, Johannes Peter Wallner, Stefan Woltran}\\ ASPARTIX-V: Utilizing Improved ASP Encodings}
\includepdf[pages=-,pagecommand={\thispagestyle{plain}}]{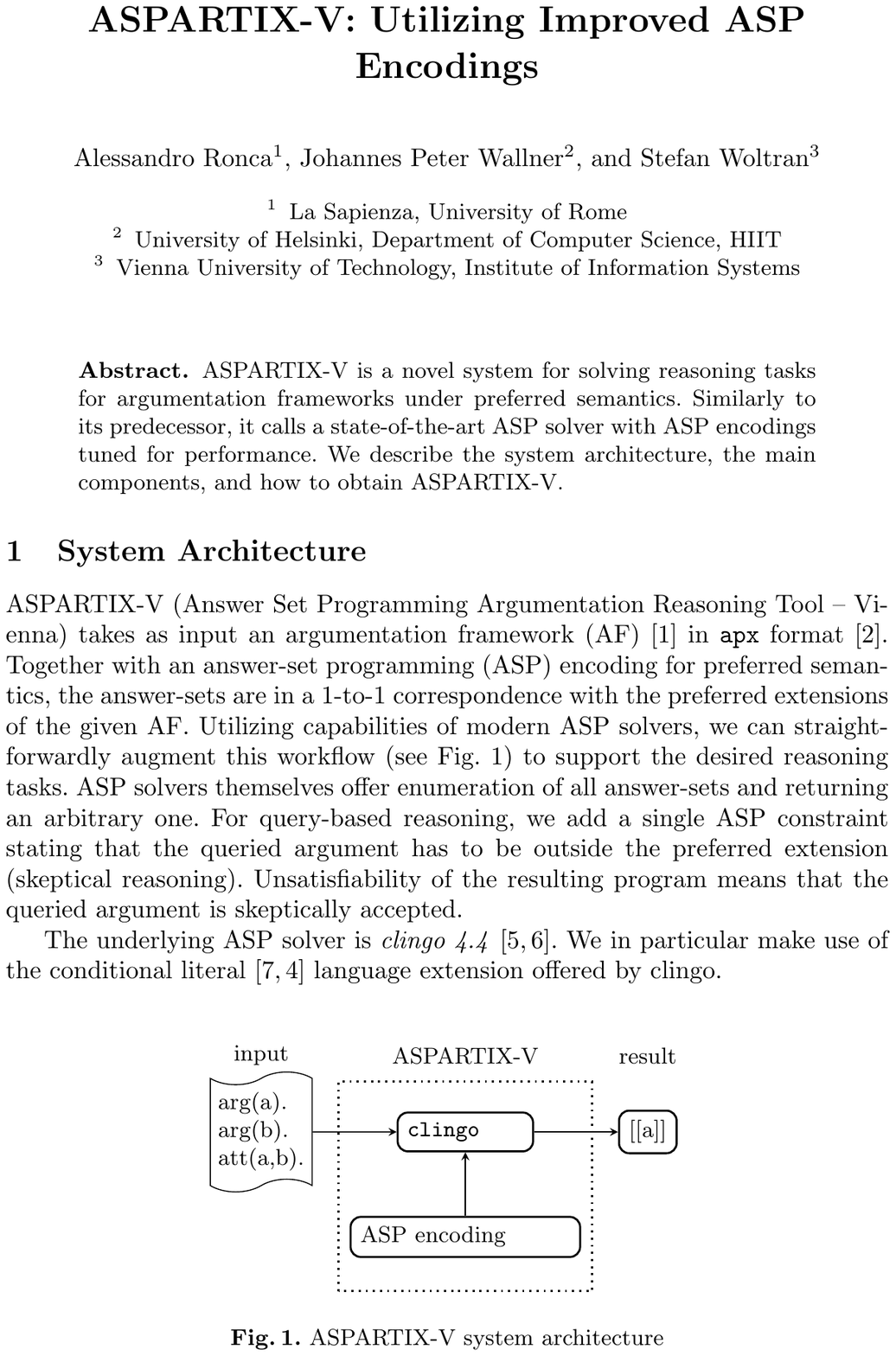}
\addcontentsline{toc}{chapter}{\normalfont \textit{Jean-Marie Lagniez, Emmanuel Lonca, Jean-Guy Mailly}\\ CoQuiAAS: Application of Constraint Programming for Abstract Argumentation}
\includepdf[pages=-,pagecommand={\thispagestyle{plain}}]{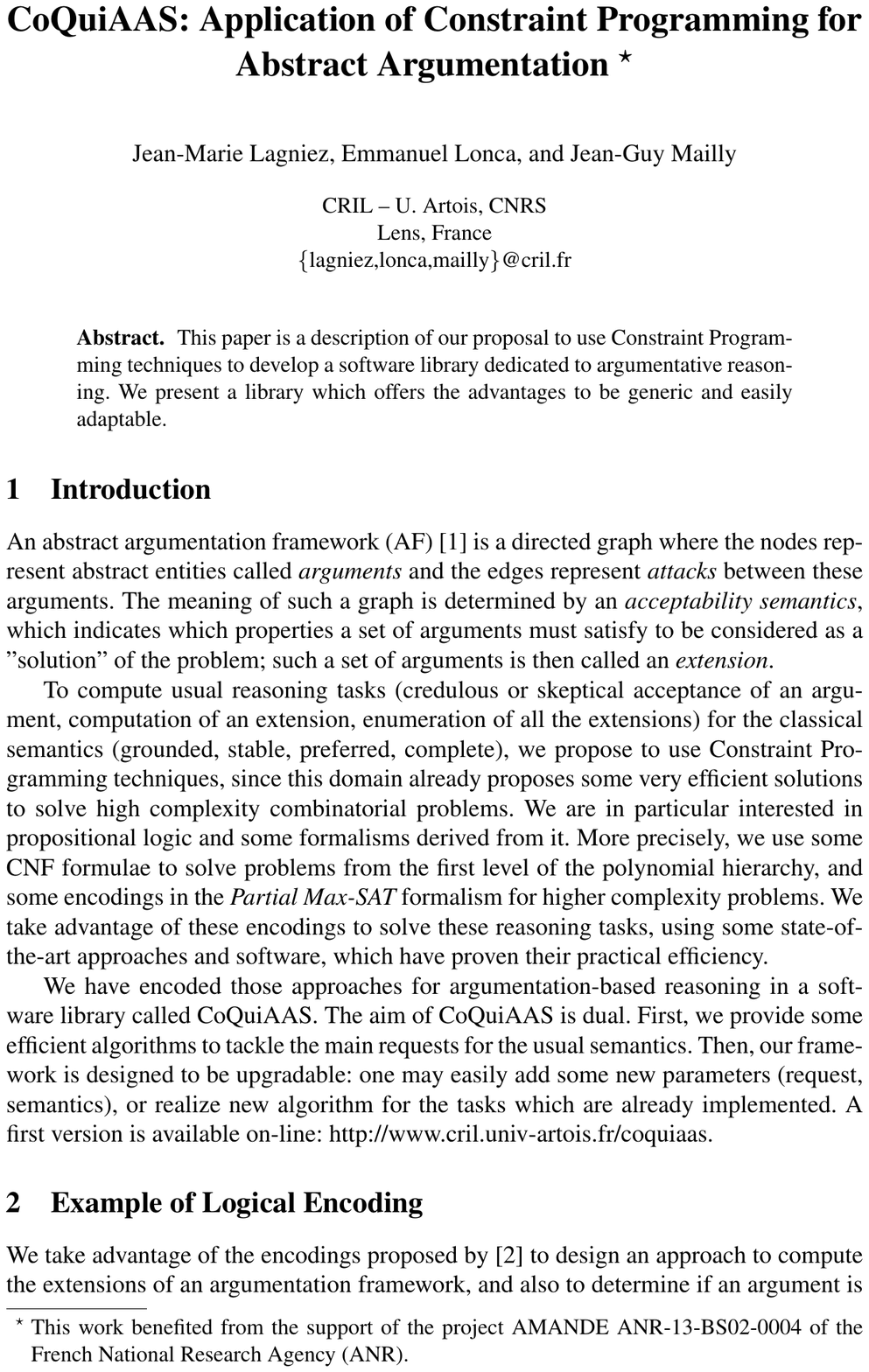}
\addcontentsline{toc}{chapter}{\normalfont \textit{Sarah Alice Gaggl, Norbert Manthey}\\ ASPARTIX-D: ASP Argumentation Reasoning Tool - Dresden}
\includepdf[pages=-,pagecommand={\thispagestyle{plain}}]{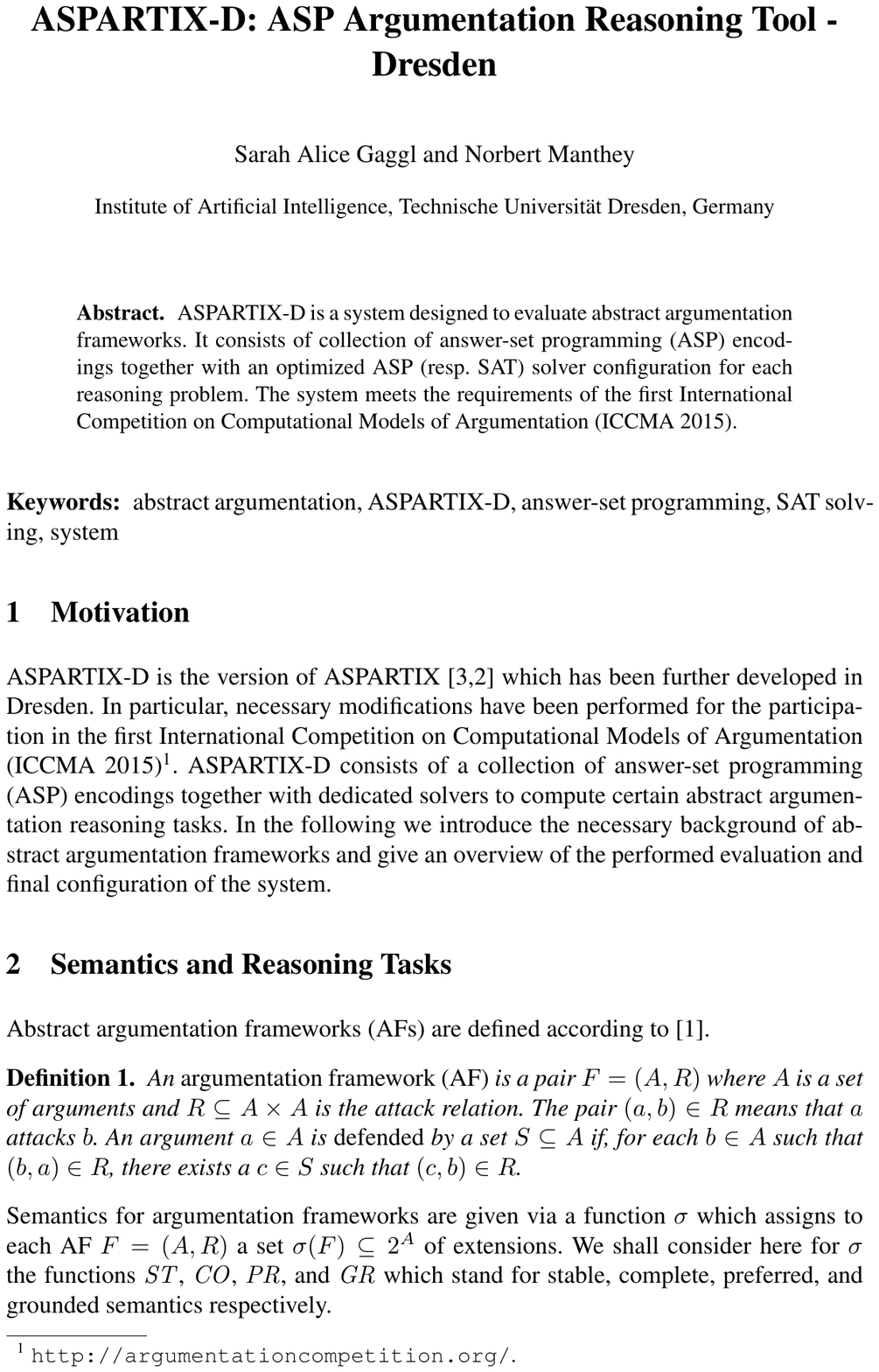}
\addcontentsline{toc}{chapter}{\normalfont \textit{Stefano Bistarelli, Fabio Rossi, Francesco Santini}\\ ConArg2: A Constraint-based Tool for Abstract Argumentation}
\includepdf[pages=-,pagecommand={\thispagestyle{plain}}]{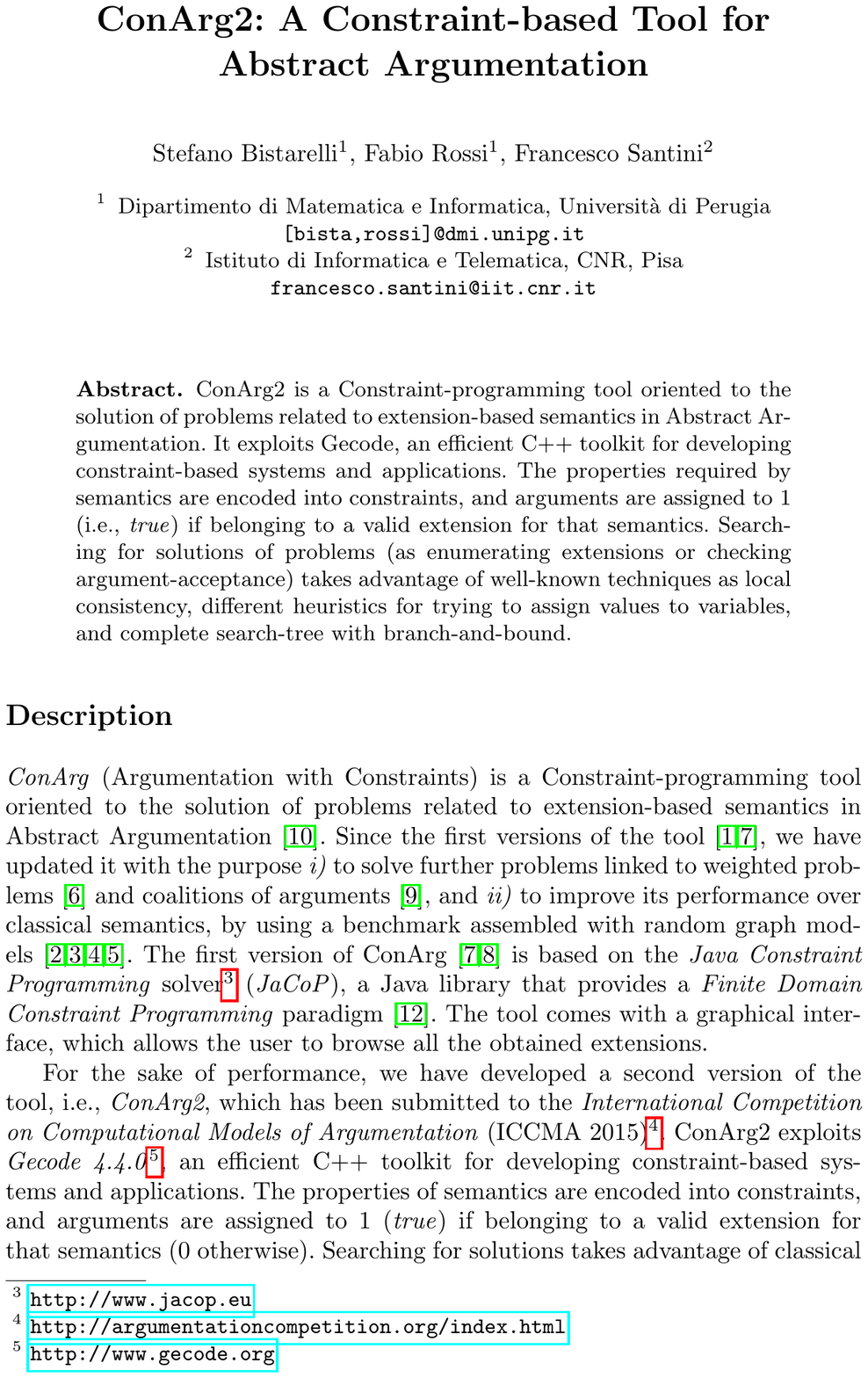}
\addcontentsline{toc}{chapter}{\normalfont \textit{Odinaldo Rodrigues}\\ GRIS: Computing traditional argumentation semantics through numerical iterations}
\includepdf[pages=-,pagecommand={\thispagestyle{plain}}]{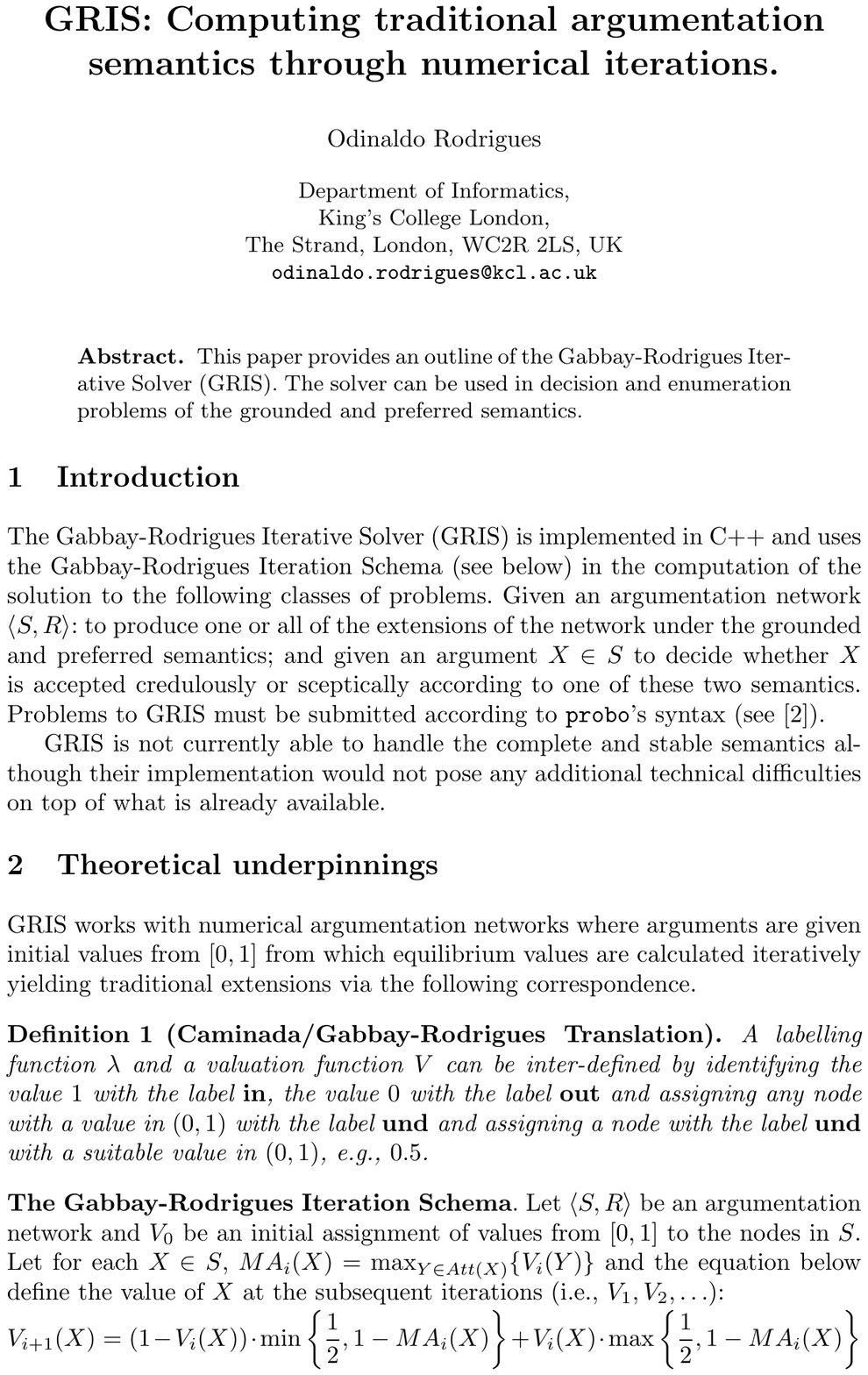}
\addcontentsline{toc}{chapter}{\normalfont \textit{Kilian Sprotte}\\ ASGL: Argumentation Semantics in Gecode and Lisp}
\includepdf[pages=-,pagecommand={\thispagestyle{plain}}]{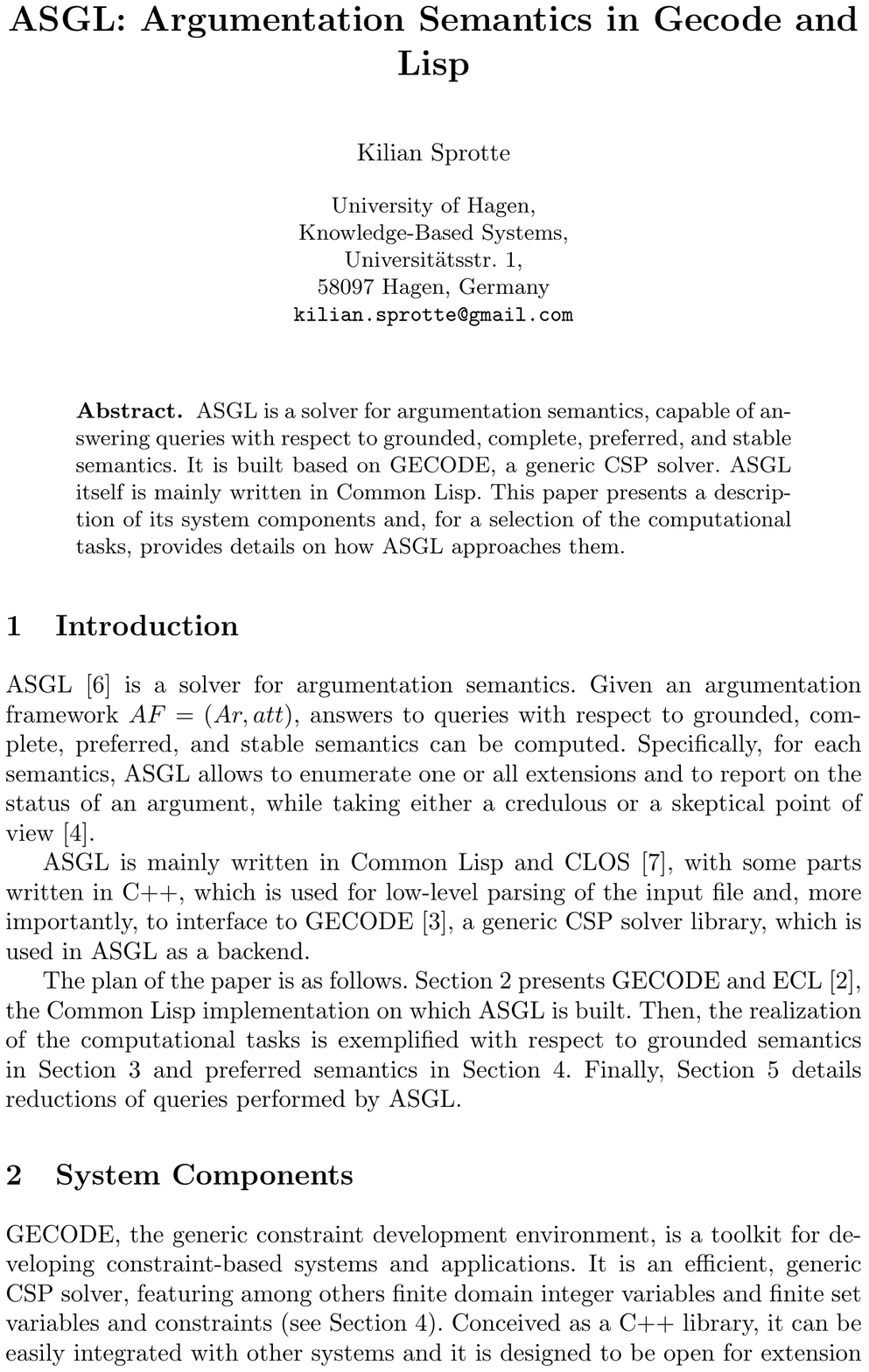}
\addcontentsline{toc}{chapter}{\normalfont \textit{Nico Lamatz}\\ LamatzSolver-v0.1: A grounded extension finder based on the Java-Collection-Framework}
\includepdf[pages=-,pagecommand={\thispagestyle{plain}}]{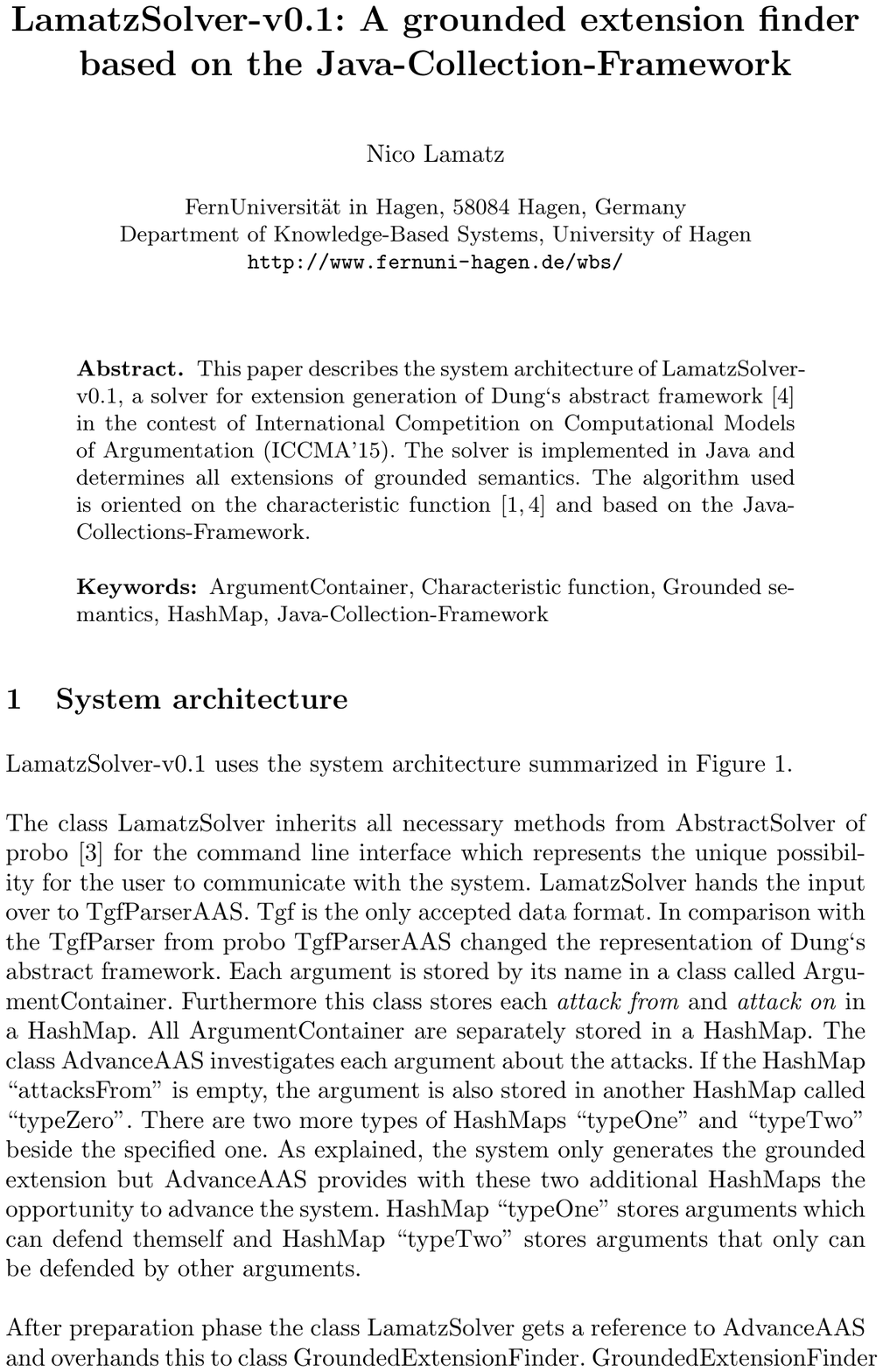}
\addcontentsline{toc}{chapter}{\normalfont \textit{Serban Groza, Adrian Groza}\\ ProGraph: towards enacting bipartite graphs for abstract argumentation frameworks}
\includepdf[pages=-,pagecommand={\thispagestyle{plain}}]{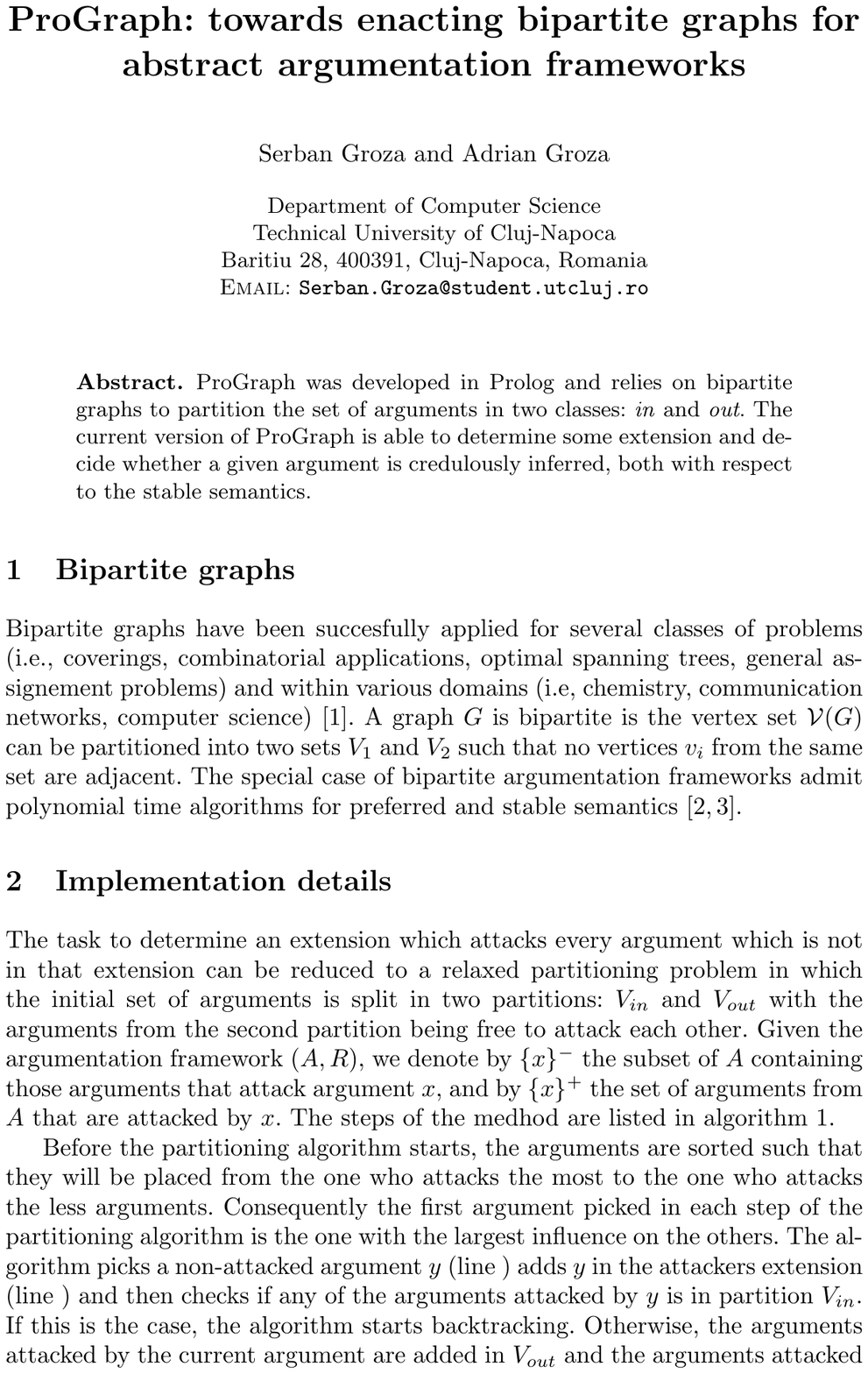}
\addcontentsline{toc}{chapter}{\normalfont \textit{Stefan Ellmauthaler and Hannes Strass}\\ DIAMOND: A System for Computing with Abstract Dialectical Frameworks}
\includepdf[pages=-,pagecommand={\thispagestyle{plain}}]{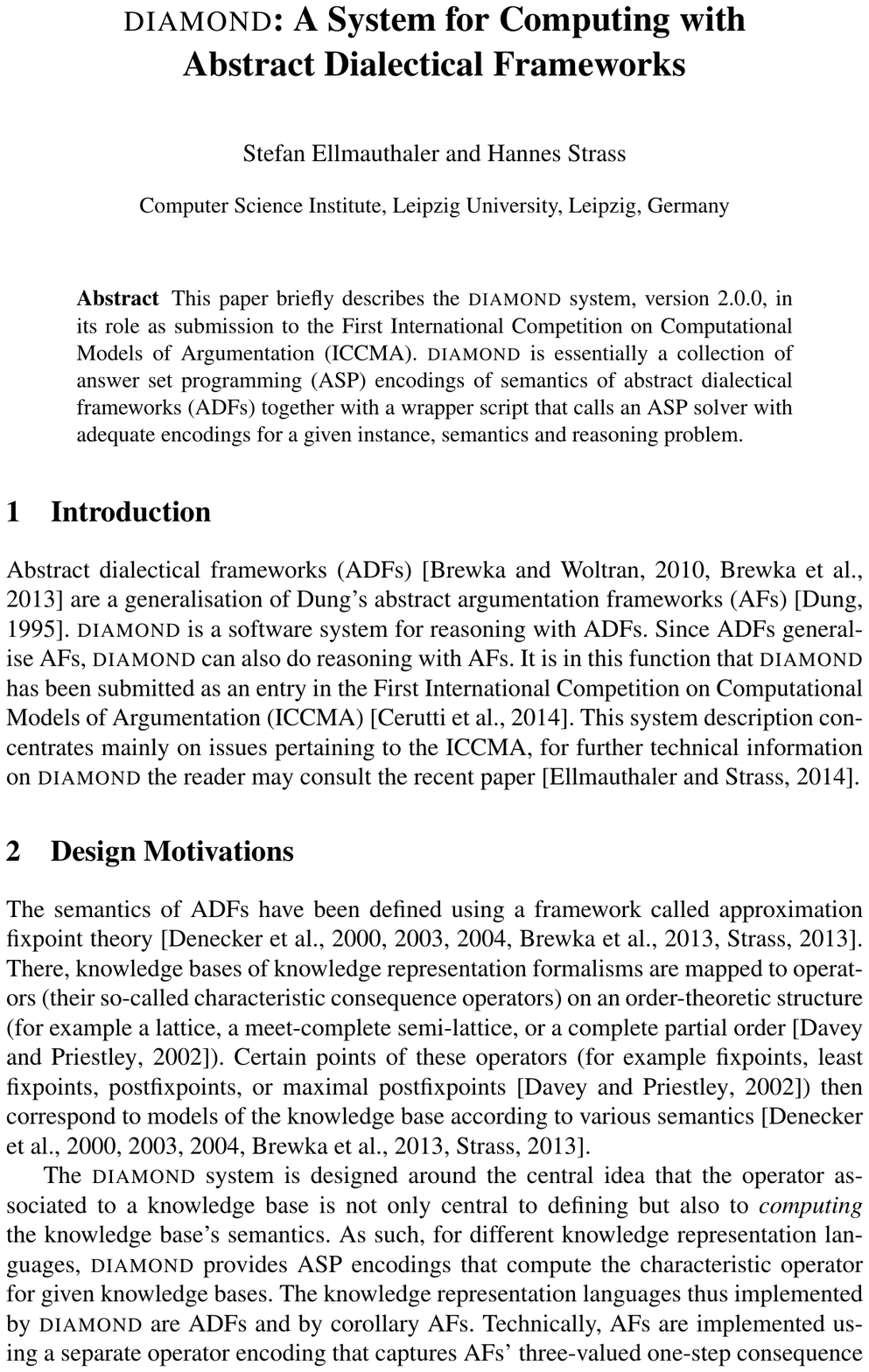}
\addcontentsline{toc}{chapter}{\normalfont \textit{Thomas F. Gordon}\\ Carneades ICCMA: A Straightforward Implementation of a Solver for Abstract Argumentation in the Go Programming Language}
\includepdf[pages=-,pagecommand={\thispagestyle{plain}}]{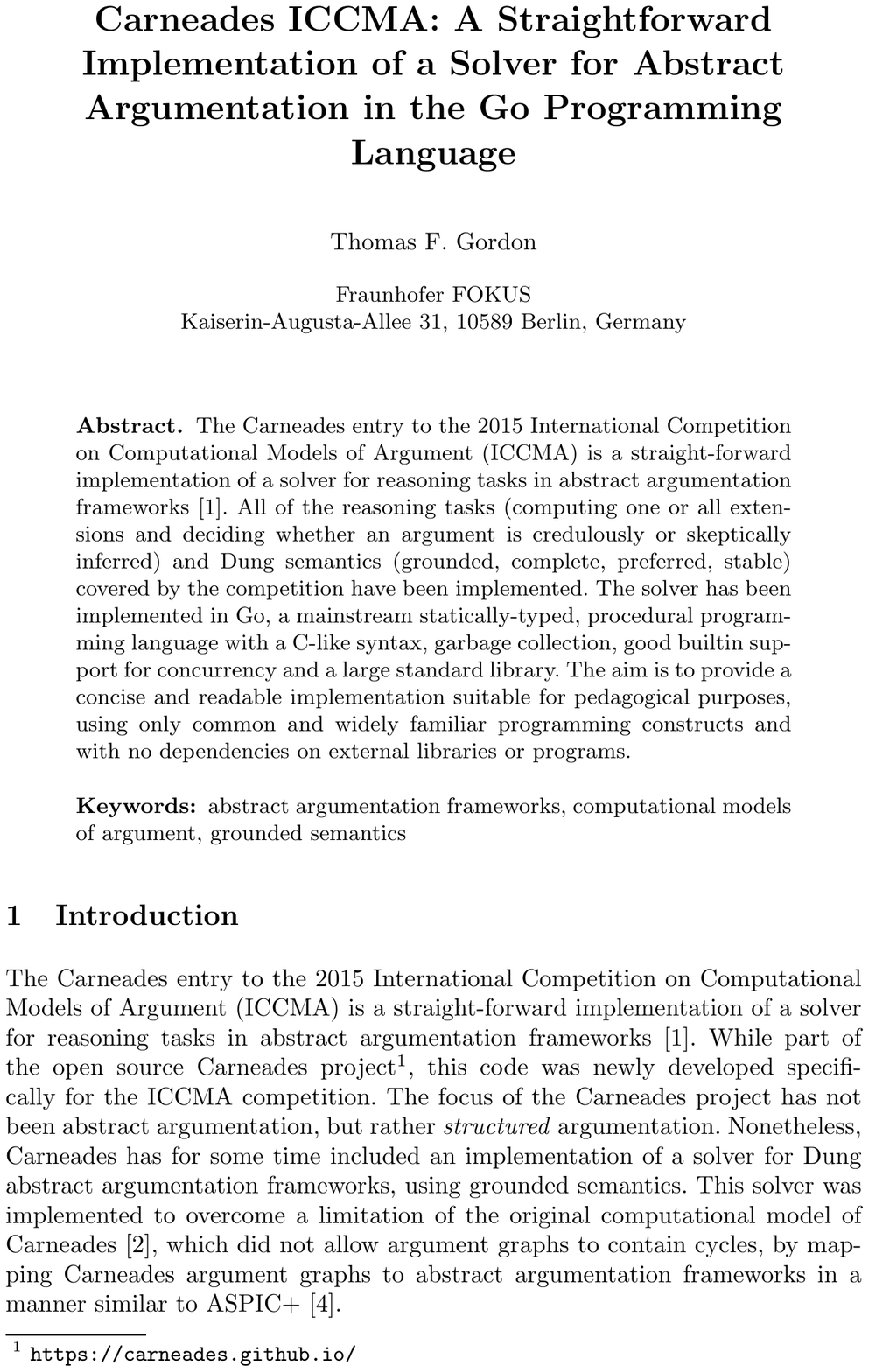}
\addcontentsline{toc}{chapter}{\normalfont \textit{Mauro Vallati, Federico Cerutti, Wolfgang Faber, Massimiliano Giacomin}\\ prefMaxSAT: Exploiting MaxSAT for Enumerating Preferred Extensions}
\includepdf[pages=-,pagecommand={\thispagestyle{plain}}]{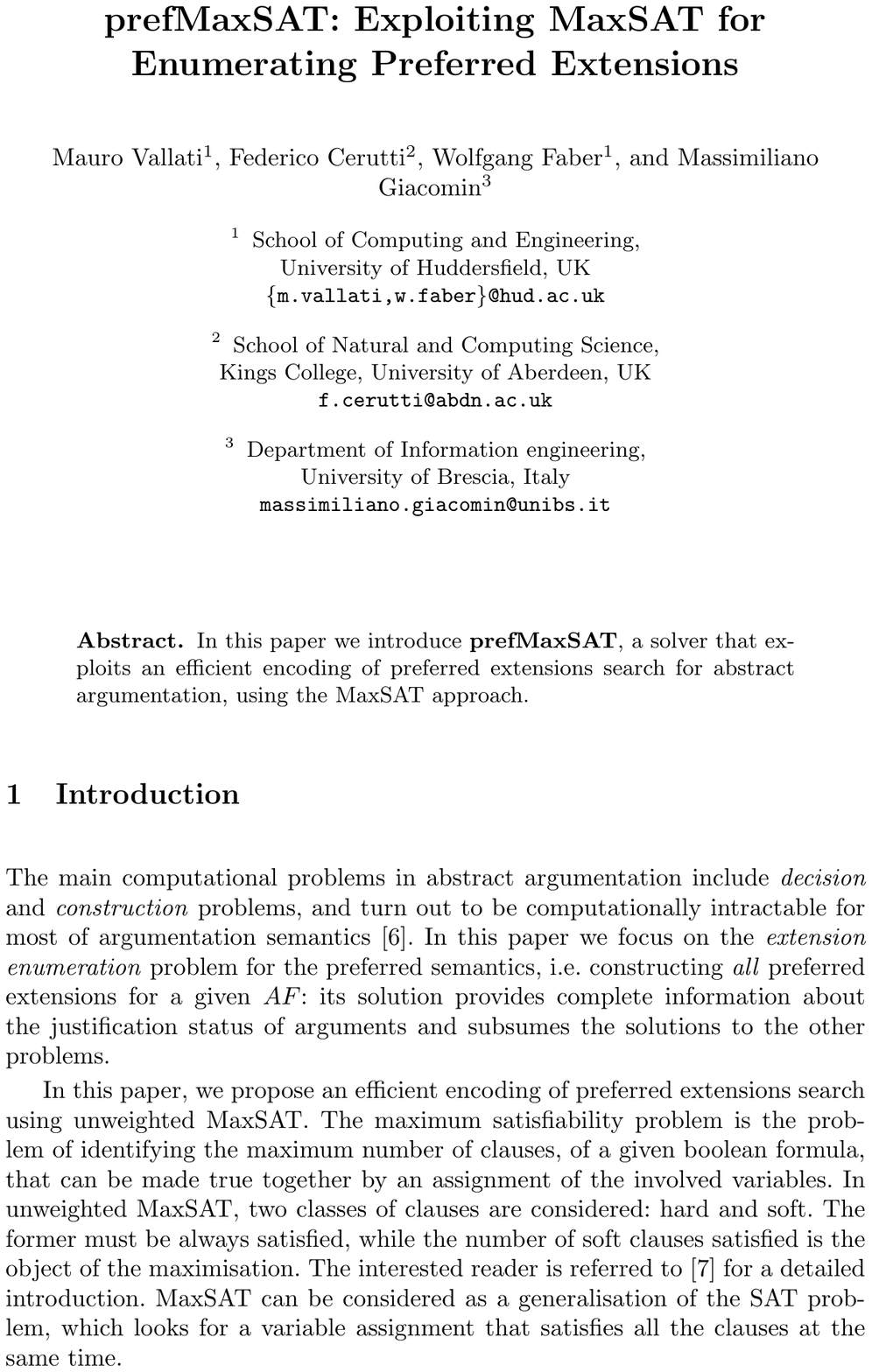}
\addcontentsline{toc}{chapter}{\normalfont \textit{Evgenios Hadjisoteriou, Michael Georgiou}\\ ASSA: Computing Stable Extensions with Matrices}
\includepdf[pages=-,pagecommand={\thispagestyle{plain}}]{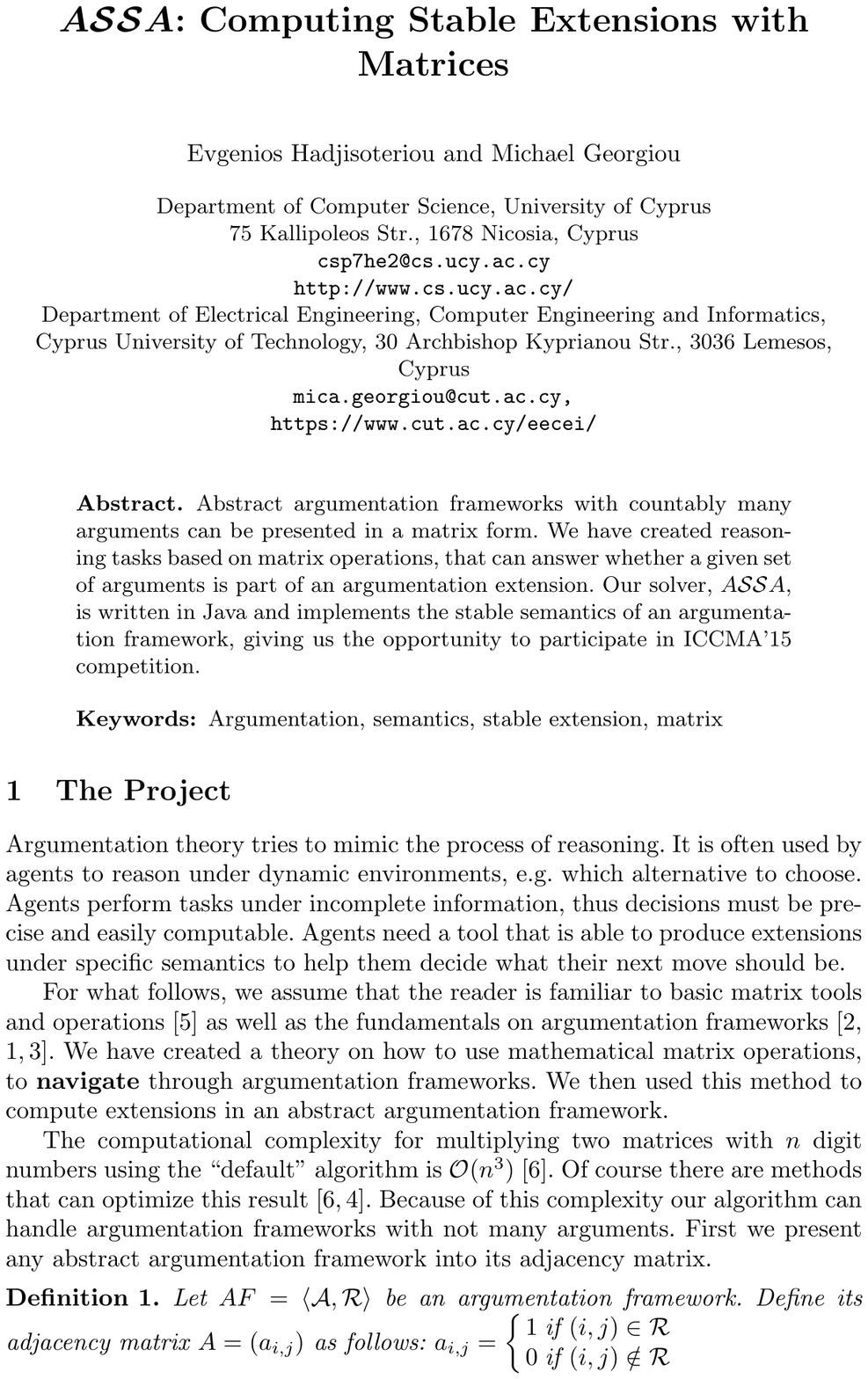}

\end{document}